\documentclass[11pt]{article}
\PassOptionsToPackage{table}{xcolor}
\usepackage[preprint]{acl}

\usepackage[utf8]{inputenc}
\usepackage[T1]{fontenc}
\usepackage{booktabs}
\usepackage{amsfonts}
\usepackage{nicefrac}
\usepackage{microtype}

\usepackage{algorithm}
\usepackage{algpseudocode}
\usepackage{caption}
\usepackage{graphicx}
\usepackage{tikz}
\usepackage{amsmath,amssymb,amsthm}
\usepackage{enumitem}
\usepackage{multirow}
\usepackage{siunitx}
\usepackage{adjustbox}
\usepackage{xcolor}
\usepackage{listings}
\graphicspath{{figures/}}
\lstdefinestyle{promptbox}{
  basicstyle=\ttfamily\scriptsize,
  breaklines=true,
  breakatwhitespace=true,
  columns=fullflexible,
  keepspaces=true,
  showstringspaces=false,
  frame=single,
  framerule=0.35pt,
  framesep=5pt,
  rulecolor=\color{black!35},
  backgroundcolor=\color{black!2},
  aboveskip=0pt,
  belowskip=0pt
}
\definecolor{pfred}{RGB}{180,30,30}
\definecolor{pfgreen}{RGB}{20,130,60}
\definecolor{pfblue}{RGB}{40,90,180}

\lstdefinestyle{promptboxcolor}{
  style=promptbox,
  moredelim={[is][\color{pfred}]{<RED>}{</RED>}},
  moredelim={[is][\color{pfgreen}]{<GREEN>}{</GREEN>}},
  moredelim={[is][\color{pfblue}]{<BLUE>}{</BLUE>}}
}
\usetikzlibrary{arrows.meta,decorations.pathreplacing}
\newcommand{\Final}{\textsc{Final}}
\newcommand{\apf}{\ensuremath{a_{\mathrm{PF}}}}
\newcommand{\PFCall}[1]{\ensuremath{\apf\!\left(#1\right)}}
\newcommand{\FinalCall}[1]{\ensuremath{\text{\Final}\!\left(#1\right)}}
\newcommand{\Tbase}{\ensuremath{\mathcal{T}_{\mathrm{base}}}}
\newcommand{\Ttrain}{\ensuremath{\mathcal{T}_{\mathrm{train}}}}
\newcommand{\Ttest}{\ensuremath{\mathcal{T}_{\mathrm{test}}}}
\newcommand{\Ind}[1]{\ensuremath{\mathbb{I}\!\left[#1\right]}}
\newcommand{\Tmax}{\ensuremath{T_{\max}}}
\newcommand{\rhopf}{\ensuremath{\rho_{\mathrm{PF}}}}

\title{CAPF: Guiding Search-Agent Rollouts with Credit-Attenuated Privileged Feedback}

\author{
Bin Chen$^{*}$ \quad Xinye Liao$^{*}$ \quad Yiming Liu$^{\dagger}$ \\
{\bfseries Xin Liao \quad Chonghan Liu} \\
\texttt{frederchen0@gmail.com}
}

\begin{document}

\maketitle
\begingroup
\renewcommand\thefootnote{}
\footnotetext{$^{*}$ Equal contribution. $^{\dagger}$ Corresponding author.}
\endgroup

\begin{abstract}
Recent LLM search agents use reinforcement learning with verifiable rewards (RLVR) to learn search-augmented reasoning from outcome rewards.
On hard problems, these agents rarely sample end-to-end successful rollouts, leaving outcome-only RLVR with few positive-reward trajectories.
We argue that improving learning on such problems requires additional guidance during training, and RLVR already contains verifier-side information that can provide it.
This information can identify errors or omissions in the agent's submitted answer and guide revision within the rollout.
We propose a training-time mechanism called \textbf{Credit-Attenuated Privileged Feedback} (CAPF), which makes this verifier-side information available through a Privileged Feedback call during training. CAPF lets the policy revise zero-reward attempts into positive-reward repair trajectories and attenuates credit for the feedback call and earlier actions to accommodate deployment without this call.
Empirical research demonstrates that CAPF improves Qwen3-4B's average exact-match score from $44.7\%$ under outcome-only RLVR to $48.5\%$ on seven open-domain QA benchmarks.
\end{abstract}

\section{Introduction}
\label{sec:introduction}

\begin{figure}[!t]
\centering
\includegraphics[width=1.0\linewidth]{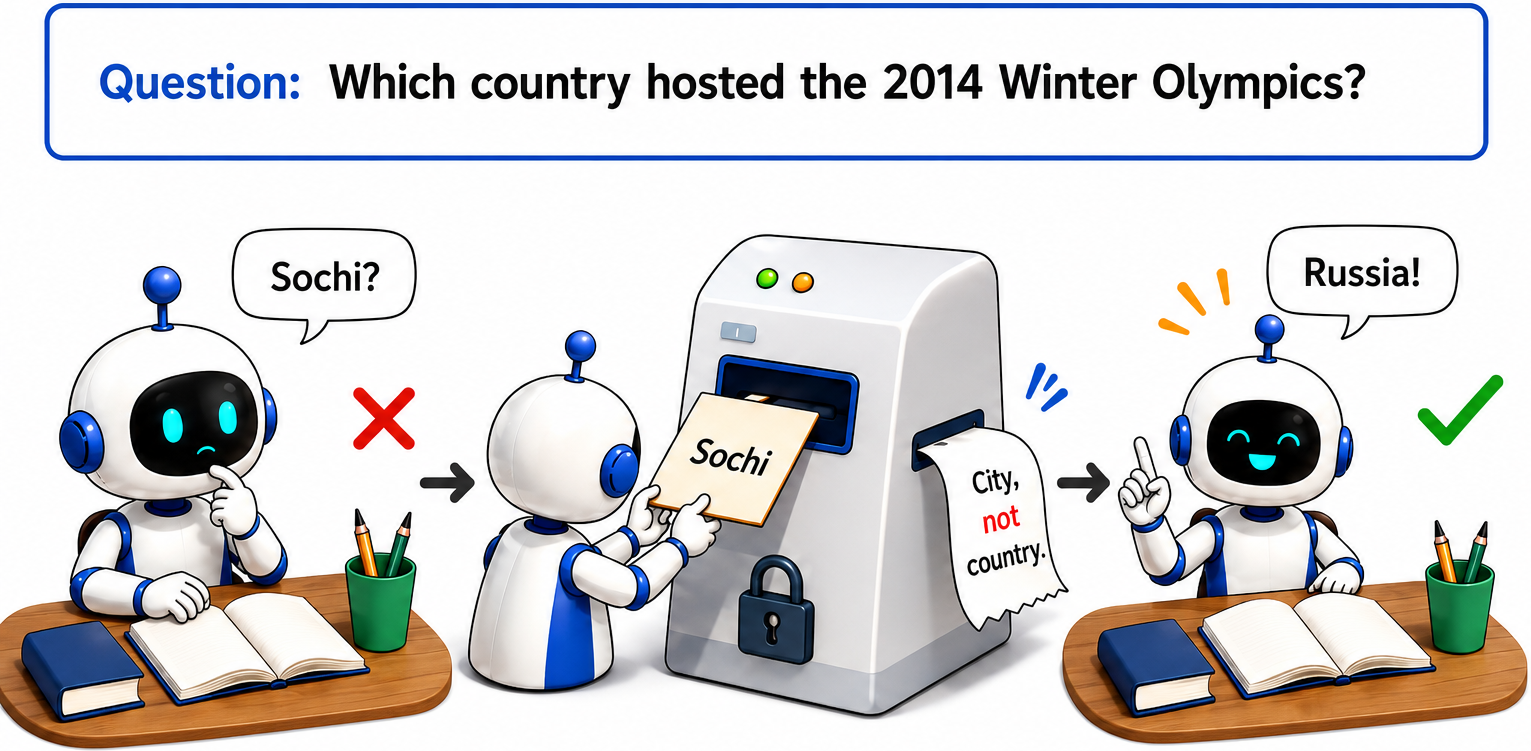}
\caption{
The agent uses Privileged Feedback to repair an uncertain candidate answer during training.
Given the question ``Which country hosted the 2014 Winter Olympics?'', the agent first drafts ``Sochi'' but remains uncertain and calls Privileged Feedback.
The feedback flags that ``Sochi'' is the host city rather than the requested country.
The agent then revises the final answer to ``Russia''.
}
\label{fig:motivation}
\vspace{-1.0ex}
\end{figure}

Search-augmented LLM agents answer open-domain questions by interleaving reasoning, search queries, retrieved evidence, and final-answer generation~\cite{nakano2021webgpt,yao2023react,li2025searcho1}.
Recent work trains such agents with reinforcement learning from verifiable rewards (RLVR), where an automatic verifier scores only the final answer~\cite{guo2025deepseek,wen2025rlvr,jin2025search,chen2025research}.
This outcome-only paradigm provides simple and scalable supervision when the agent samples enough successful rollouts.
On hard problems, however, search agents rarely reach the correct answer within the rollout budget, so most attempts receive zero reward.
Outcome-only RLVR therefore obtains little positive learning signal from the problems where additional guidance would be most useful.

The verifier-side information behind verifiable rewards can help recover this missing signal during training.
Because the training environment has access to reference answers, it can compare the agent's submitted answer with the references and return a short feedback message.
The feedback can point out what is wrong or missing, such as an incorrect answer type, a missing constraint, or unsupported evidence.
This gives the agent a chance to revise within the same rollout instead of ending the attempt with zero reward.

We call this verifier-generated feedback \textbf{Privileged Feedback}.
Figure~\ref{fig:motivation} illustrates how this feedback can turn a zero-reward attempt into a repair trajectory.
The agent first submits ``Sochi'' for a question that asks for the host country of the 2014 Winter Olympics.
Privileged Feedback identifies the answer-type error and allows the agent to revise within the rollout.
These repaired trajectories provide learning signal that outcome-only RLVR would miss, but they are not ordinary end-to-end successes.
Their post-feedback suffixes may contain useful repair behavior, while their final success passes through privileged information that will not be available at deployment.
The question is how to use these successes as learning signal while preventing the policy from relying on Privileged Feedback.

We propose \textbf{Credit-Attenuated Privileged Feedback} (CAPF) to address this problem.
CAPF adds a Privileged Feedback call to the training action space and removes this call at deployment.
It computes turn-level returns with a Privileged Feedback retention factor.
When credit propagates backward across a Privileged Feedback call, CAPF attenuates the return for the call and earlier actions while preserving full credit for actions after the call.
This gives full credit to post-feedback repair actions while giving weaker credit to the feedback call and earlier actions.
Evaluated without Privileged Feedback, CAPF improves Qwen3-4B's average exact-match score on seven open-domain QA benchmarks from $44.7\%$ under outcome-only RLVR to $48.5\%$.

Our main contributions are as follows:
\begin{enumerate}[leftmargin=*, itemsep=0.1pt]
\item We introduce \textbf{Privileged Feedback} for search-agent RLVR, using verifier-side information during training to create repair trajectories from attempts that would otherwise receive zero reward.
\item We propose \textbf{Credit-Attenuated Privileged Feedback} (CAPF), which adds a Privileged Feedback call to the training action space and attenuates return when credit crosses such calls.
\item We show that CAPF improves deployed search-agent performance after Privileged Feedback is removed.
\end{enumerate}

\section{Related Work}
\label{sec:related_work}

CAPF builds on search-agent RLVR but uses verifier-side feedback to recover learning signal from otherwise zero-reward attempts.
We compare this design with prior work on additional supervision for sparse rewards, learning with privileged information, and reward shaping.

\paragraph{Search-agent reinforcement learning.}
Search-augmented language agents interleave reasoning with actions that retrieve and use external evidence~\cite{nakano2021webgpt,yao2023react,li2025searcho1}.
Recent RLVR systems optimize this interaction loop from final correctness, improving query selection, evidence use, and long-horizon search behavior~\cite{jin2025search,chen2025research,zheng2025stepsearch,gao2025beyond,zheng2025deepresearcher,lu2026searchselfplay}.
These systems train policies to use actions that remain available at test, such as search, reasoning, and final-answer generation.
CAPF instead augments the training action space with a Privileged Feedback call.
When the policy issues this call, the environment returns a feedback message as an observation.
At deployment, the call is removed.
This shifts the question from learning to use a persistent search action to learning from repair trajectories produced by training-time calls.

\paragraph{Additional supervision for sparse rewards.}
One line of work addresses sparse outcome rewards by adding supervision beyond final correctness.
Process supervision and step-level verification score intermediate reasoning steps~\cite{uesato2022process,lightman2023let}, while search-agent methods add denser rewards~\cite{luo2025infoflow}, turn-level feedback~\cite{xie2026tips}, or instructive feedback~\cite{li2025reseek,shi2025autorefine} to guide exploration and evidence use.
These methods can make sparse-reward learning easier, but they often require additional annotations, reward design, or auxiliary evaluators.
CAPF instead reuses verifier-side information already present in RLVR to create repair trajectories without additional process annotations.

\paragraph{Learning with privileged information.}
Learning with privileged information studies settings where training examples contain extra information unavailable at test time~\cite{vapnik2009lupi,pechyony2010lupi}.
Recent LLM work similarly uses privileged teacher feedback, oracle guidance, or verified traces to improve exploration, supervision, or distillation~\cite{choudhury2025leap,qu2026pope}.
CAPF also falls within this setting, but does not use privileged information as an external teacher signal or extra label.
Instead, CAPF lets the policy issue Privileged Feedback calls inside the rollout rather than supplying privileged information as an external training signal. Each call is recorded in the trajectory and can be accounted for in credit assignment.

\paragraph{Reward shaping and credit assignment.}
Reward shaping modifies the reward function to guide learning~\cite{ng1999policy}, and transition-dependent objectives study returns that depend on transition structure~\cite{white2017unifying,pitis2019rethinking}.
CAPF is related because it changes how final success is assigned to actions in a trajectory.
It differs by leaving the final-answer verifier reward unchanged.
Rather than adding dense rewards or imposing a hand-coded cost for using Privileged Feedback, CAPF attenuates return only when final success propagates backward across a Privileged Feedback call.
This keeps repaired trajectories useful for learning while reducing credit to behavior before the training-time feedback.

\section{Method}
\label{sec:method}

\begin{figure*}[!t]
\centering
\includegraphics[width=0.98\textwidth]{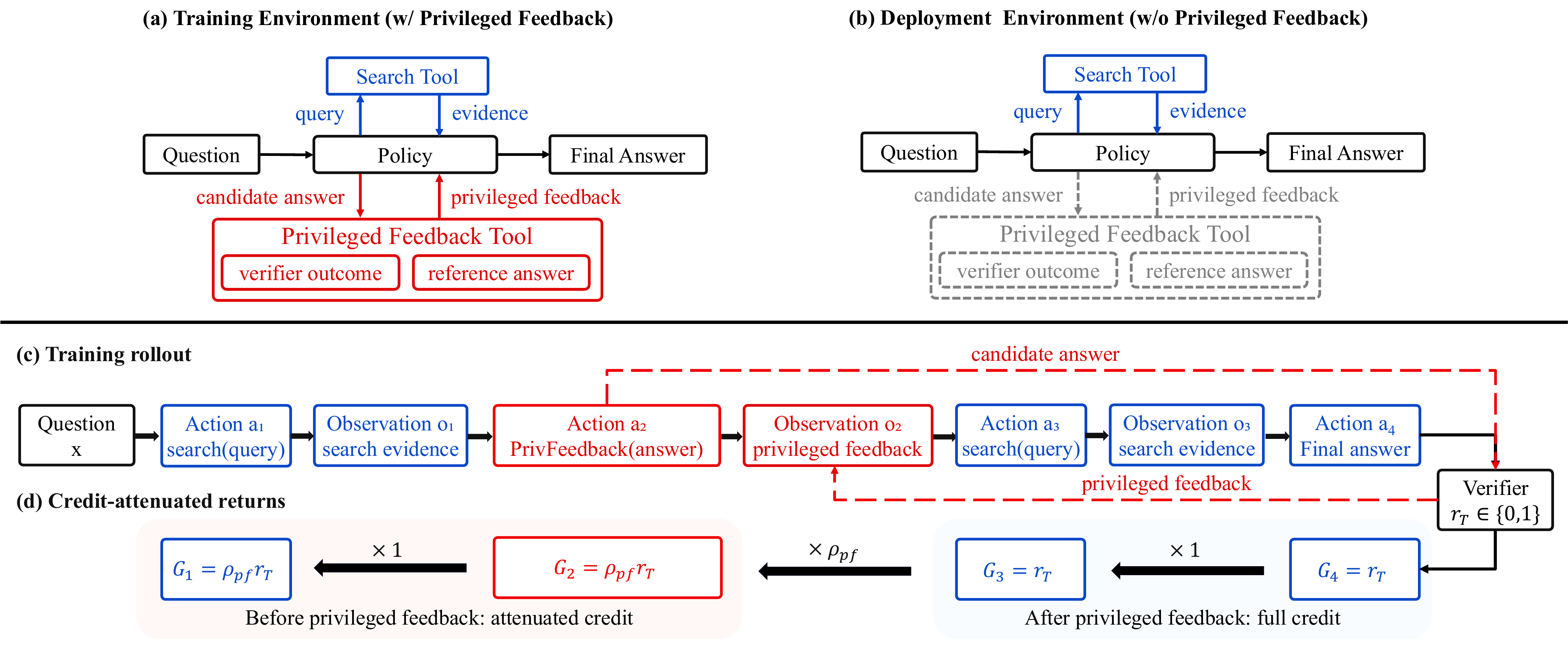}
\caption{
CAPF adds a Privileged Feedback call to the training action space and removes this call at deployment.
The final-answer verifier reward remains unchanged.
CAPF attenuates return only when final success propagates backward across a Privileged Feedback call.
(a) Training environment with Privileged Feedback;
(b) deployment environment without Privileged Feedback;
(c) training rollout with Privileged Feedback;
(d) credit-attenuated returns.}
\label{fig:capf_overview}
\vspace{-1.0ex}
\end{figure*}

CAPF adds a Privileged Feedback call to the training action space and removes this call at deployment.
When the policy issues this call with a candidate answer, the environment returns a feedback message as an observation.
The policy can then revise within the rollout, turning some otherwise zero-reward attempts into positive-reward repair trajectories.
Since each Privileged Feedback call is visible in the trajectory, CAPF can attenuate return for the call and earlier actions while preserving full credit for post-feedback actions.

\subsection{Search-Agent RLVR with Deployment Constraints}
\label{sec:method_setting}

Each instance contains a question $x$ and a reference answer $y^\star$.
At each nonterminal step, the policy selects an action $a_t$.
The search environment executes this action and returns an observation $o_t$.
Thus, the policy conditions on the history $h_t=(x,a_{<t},o_{<t})$.

The base action space $\Tbase$ contains ordinary search-agent actions and a final-answer action $\FinalCall{y}$.
When the policy emits $\FinalCall{y_T}$ at step $T$, the episode terminates.
An exact-answer verifier then compares $y_T$ with $y^\star$ and returns:
\begin{equation}
\label{eq:terminal_reward}
R(\tau)=\mathsf{V}(y_T,y^\star)\in\{0,1\}.
\end{equation}
All nonterminal task rewards are zero.
When the policy does not emit a final answer within the rollout budget, we set $R(\tau)=0$.

Training and deployment differ by one action:
\begin{equation}
\label{eq:train_test_actions}
\Ttrain=\Tbase\cup\{\apf\},
\quad
\Ttest=\Tbase
\end{equation}
The augmented action $\apf$ is used only during training rollout collection.
All reported benchmark results use $\Ttest$, with the Privileged Feedback call absent from the action space.

\subsection{Privileged Feedback Call}
\label{sec:method_feedback}

A Privileged Feedback call has the form:
\begin{equation}
\label{eq:feedback_action}
a_t=\PFCall{\hat{y}_t},
\end{equation}
where $\hat{y}_t$ denotes the policy's current candidate answer.
This nonterminal action does not directly alter the task reward.
Instead, the environment compares $\hat{y}_t$ with the reference answer $y^\star$, obtains the verifier label $\mathsf{V}(\hat{y}_t,y^\star)$, and appends a concise feedback message as an environment observation.
Subsequent policy actions condition on the updated history.

This rollout structure separates policy text from environment text.
The policy produces search actions, Privileged Feedback calls, and final answers.
The environment returns retrieval results and feedback messages.
Tokens returned by the environment are masked and therefore do not contribute to the policy log probability or the training loss.
The reference answer and verifier label are used only inside the environment-side feedback generator.
Since the feedback message relies on verifier-side information unavailable at deployment, we still regard it as privileged information, even though prompt constraints and exact string redaction limit direct copying from the reference answer.

\subsection{Credit-Attenuated Privileged Feedback}
\label{sec:method_capf}

Successful repair trajectories can recover learning signal from attempts that outcome-only RLVR would score as zero.
Their post-feedback suffixes may contain transferable behavior, such as issuing a more effective query, inspecting more relevant evidence, or correcting the answer type.
The feedback call and earlier actions should not receive the same credit as unassisted successful behavior, since final success has passed through an environment observation unavailable at deployment.
Ordinary RLVR does not make this distinction and assigns the same final reward signal to all policy actions in a successful trajectory.

CAPF implements this distinction by attenuating return only when backward credit crosses a Privileged Feedback call.
Let $T$ be the index of the final-answer action.
For a policy decision at step $t$, define
\begin{equation}
N_t^{\mathrm{PF}}(\tau)
=
\sum_{s=t}^{T-1}
\Ind{a_s=\PFCall{\cdot}} ,
\end{equation}
as the number of Privileged Feedback calls from step $t$ through step $T-1$, including $a_t$ itself when it is a Privileged Feedback call.
By convention, $N_T^{\mathrm{PF}}(\tau)=0$.

For a retention factor $0<\rhopf\le 1$, CAPF defines the return for the action taken at step $t$ as
\begin{equation}
\label{eq:capf_return}
G_t^{\mathrm{CAPF}}(\tau)
=
R(\tau)\,\rhopf^{N_t^{\mathrm{PF}}(\tau)}.
\end{equation}
Thus, actions after the last Privileged Feedback call receive the full verifier reward.
The feedback call and earlier actions receive attenuated credit.
If one Privileged Feedback call lies between action $a_t$ and the final answer, including $a_t$ itself, the action receives $\rhopf R(\tau)$.
If $k$ such calls lie between $a_t$ and the final answer, the action receives $\rhopf^k R(\tau)$.

\subsection{CAPF Policy Gradient Objective}
\label{sec:method_optimization}

We optimize the policy with \textsc{REINFORCE++}~\cite{williams1992simple,sutton1999policy,hu2025reinforcepp}.
CAPF modifies only the policy-gradient target and keeps the optimizer, rollout budgets, final-answer verifier, baseline subtraction, and advantage normalization the same as in the outcome-only run.
For each policy action, CAPF replaces the ordinary return based on $R(\tau)$ with $G_t^{\mathrm{CAPF}}$ from Eq.~\eqref{eq:capf_return}.

The return targets $G_t^{\mathrm{CAPF}}$ can be computed with a single backward pass over the trajectory.
Starting from the reward from the final-answer verifier, the pass multiplies the running return by $\rhopf$ each time it crosses a Privileged Feedback call.
Algorithm~\ref{alg:capf_return} summarizes this computation.

\begin{algorithm}[t]
\small
\caption{CAPF return computation}
\label{alg:capf_return}
\begin{algorithmic}[1]
  \Require Trajectory actions $a_0,\ldots,a_T$, final-answer verifier reward $R(\tau)$, retention factor $\rhopf$
  \State $g \gets R(\tau)$
  \State $G_T \gets g$
  \For{$t=T-1,\ldots,0$}
    \If{$a_t=\PFCall{\cdot}$}
      \State $g \gets \rhopf \cdot g$
    \EndIf
    \State $G_t \gets g$
  \EndFor
  \State \Return $G_0,\ldots,G_T$
\end{algorithmic}
\vspace{-1.0ex}
\end{algorithm}

Let $\mathcal{A}_{\pi}(\tau)$ denote the set of policy-generated action indices in the trajectory.
For text actions, $\log \pi_\theta(a_t\mid h_t)$ is the sum of token log probabilities over the generated action span.
The policy-gradient loss is
\begin{equation}
\label{eq:policy_loss}
\mathcal{L}_{\mathrm{pg}}(\theta)
=
-\sum_{t\in \mathcal{A}_{\pi}(\tau)}
\widehat{A}_t
\log \pi_\theta(a_t\mid h_t).
\end{equation}
The advantage $\widehat{A}_t$ is computed from $G_t^{\mathrm{CAPF}}$ using the same baseline subtraction and advantage normalization as the outcome-only run.
Environment-generated text, including retrieval results and feedback messages, is outside $\mathcal{A}_{\pi}(\tau)$.
It is not assigned log probability and is masked from the loss.
Malformed actions and parser failures follow the same environment rules in outcome-only RL and CAPF.
Appendix~\ref{app:capf_impl} gives the prompts, tool schema for the Privileged Feedback call, redaction rule, and rollout traces.

\subsection{Objective Decomposition}
\label{sec:method_decomposition}

To analyze how the CAPF objective uses Privileged Feedback, we decompose trajectories by the number of Privileged Feedback calls.
Let $S_k(\pi)$ denote the probability that the policy succeeds after using Privileged Feedback exactly $k$ times:
\begin{equation}
S_k(\pi)
=
\Pr_{\tau\sim p_{\Ttrain}^{\pi}}
\left[
R(\tau)=1,\,
N_0^{\mathrm{PF}}(\tau)=k
\right]
\end{equation}
The CAPF-weighted start-state success can then be written as
\begin{equation}
\label{eq:objective_decomp}
\begin{aligned}
J_{\rhopf}^{\mathrm{start}}(\pi)
&=
\sum_{k\ge 0}\rhopf^k S_k(\pi) \\
&=
S_0(\pi)
+
\underbrace{
\sum_{k\ge 1}\rhopf^k S_k(\pi)
}_{B_{\rhopf}(\pi)} .
\end{aligned}
\end{equation}
Here, $S_0$ is success without Privileged Feedback, while $B_{\rhopf}$ is the attenuated contribution from feedback-assisted successes.

This decomposition is only an analysis tool for measuring how Privileged Feedback contributes to the learning signal during training.
The terms with $k\ge 1$ depend on trajectories that use Privileged Feedback and cannot be replayed unchanged under $\Ttest$.
CAPF still keeps these trajectories because their post-feedback suffixes may contain repair behaviors that transfer to no-feedback execution.

\section{Experiments}
\label{sec:experiments}

\begin{table*}[!t]
\centering
\caption{Deployment evaluation with \texttt{Qwen3-4B-Thinking-2507} after removing the Privileged Feedback call.
The upper block gives the main attribution comparison; the lower block reports retention factor sensitivity.
All scores are exact-match (EM) values in \%; Avg. is the benchmark-level macro-average.}
\label{tab:qa_matched}
\smallskip
\begingroup
\small
\setlength{\tabcolsep}{3.6pt}
\begin{tabular}{lcccccccc}
\toprule
\multirow{2}{*}{Method}
& \multicolumn{3}{c}{\textbf{Single-hop QA}}
& \multicolumn{4}{c}{\textbf{Multi-hop QA}}
& \multirow{2}{*}{\textbf{Avg.}} \\
\cmidrule(lr){2-4} \cmidrule(lr){5-8}
& NQ & TriviaQA & PopQA
& HotpotQA & 2Wiki & MuSiQue & Bamboogle & \\
\midrule
Direct inference
  & 35.5 & 64.4 & 37.7
  & 33.2 & 37.6 & 8.3 & 32.8 & 35.6 \\
Outcome-only RL
  & 40.5 & 65.8 & 40.4
  & 43.0 & 51.4 & 19.3 & 52.8 & 44.7 \\
CAPF ($\rhopf=1.0$)
  & 41.4 & 67.8 & 42.5
  & 45.7 & 49.5 & 21.9 & 53.6 & 46.1 \\
CAPF ($\rhopf=0.8$)
  & \textbf{43.6} & \textbf{69.4} & \textbf{44.5}
  & \textbf{48.5} & 53.6 & \textbf{25.2} & \textbf{54.4} & \textbf{48.5} \\
\midrule
\multicolumn{9}{l}{\emph{CAPF retention factor sensitivity}} \\
CAPF ($\rhopf=0.5$)
  & 41.6 & 67.9 & 43.5
  & 46.8 & \textbf{53.9} & 23.2 & \textbf{54.4} & 47.3 \\
CAPF ($\rhopf=0.6$)
  & 42.5 & 68.2 & 44.0
  & 48.0 & 53.6 & 23.9 & 53.6 & 47.7 \\
\bottomrule
\end{tabular}
\endgroup
\vspace{-1.0ex}
\end{table*}

We evaluate CAPF under deployment constraints.
All reported benchmark results are measured after removing the Privileged Feedback call.
Training-environment probes are used only to analyze how repair-based learning transfers to deployment execution.

\paragraph{Benchmarks.}
We evaluate final-answer exact match (EM) on seven open-domain QA benchmarks: NQ~\cite{kwiatkowski2019natural}, TriviaQA~\cite{joshi2017triviaqa}, PopQA~\cite{mallen2022whennottrust}, HotpotQA~\cite{yang2018hotpotqa}, 2WikiMultiHopQA~\cite{ho2020twikimultihopqa}, MuSiQue~\cite{trivedi2021musique}, and Bamboogle~\cite{press2022compositionalitygap}.
Avg. denotes the benchmark-level macro-average.
Appendix~\ref{app:data_eval} gives the splits, example counts, and answer parsing details.

\paragraph{Training setup.}
All 4B attribution runs train \texttt{Qwen3-4B-Thinking-2507}~\cite{qwen2025qwen3} with OpenRLHF \texttt{reinforce++}~\cite{hu2024openrlhf,hu2025reinforcepp} on the same training data used by Search-R1~\cite{jin2025search,jin2024flashrag}.
The feedback generator is a fixed \texttt{Qwen3-30B-A3B-Instruct-2507}.
Relative to outcome-only RL, CAPF adds the Privileged Feedback call and feedback instructions during training, and uses the return in Eq.~\eqref{eq:capf_return}.
The deployed policy cannot call Privileged Feedback.
Appendix~\ref{app:training_protocol} provides the retrieval, optimization, and rollout details.

\subsection{CAPF Improves Deployment Performance}
\label{sec:exp_main}

Table~\ref{tab:qa_matched} reports deployment performance after removing the Privileged Feedback call from the action space.
Outcome-only RL serves as a strong baseline relative to direct inference, increasing average EM from $35.6\%$ to $44.7\%$.
CAPF with $\rhopf=0.8$ reaches $48.5\%$, exceeding this baseline by $+3.8$ points and the unattenuated Privileged Feedback variant by $+2.4$ points.
The improvement is consistent across benchmark families, with average EM rising from $48.9\%$ to $52.5\%$ on single-hop tasks and from $41.6\%$ to $45.4\%$ on multi-hop tasks.
The remaining CAPF rows in Table~\ref{tab:qa_matched} vary the retention factor and are analyzed in Section~\ref{sec:exp_rho}.

\subsection{Feedback-Assisted Repairs Transfer to No-Feedback Execution}
\label{sec:exp_transfer}

\begin{figure}[t]
\centering
\includegraphics[width=\linewidth]{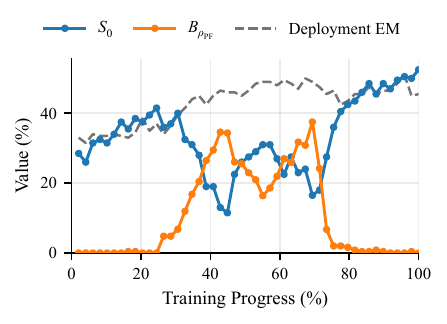}
\caption{Mechanism probe for CAPF with $\rhopf=0.8$ on a fixed 200-example set.
Training-environment success is decomposed into $S_0$ and $B_{\rhopf}$ from Eq.~\eqref{eq:objective_decomp}.
Deployment EM is measured on the same probe without Privileged Feedback.}
\label{fig:source_of_gain}
\vspace{-1.0ex}
\end{figure}

Benchmark results show that CAPF improves deployment performance, but they do not explain where the gain comes from.
We use a fixed 200-example probe to relate training-environment success to deployment execution.
We sample the probe from the RL training data before training, keep it fixed across runs, and use it only for analysis.
At each checkpoint, we evaluate the same policy in the training environment, where the Privileged Feedback call is available, and in the deployment environment, where this call is removed.
We report $S_0$ for successes without any feedback call and compute $B_{\rhopf}=\sum_{k\ge 1}\rhopf^k S_k$ for attenuated feedback-assisted successes.
Each $S_k$ is estimated from rollout counts on the training-environment probe.

For CAPF with $\rhopf=0.8$, Figure~\ref{fig:source_of_gain} shows an early dependence on Privileged Feedback.
At step 330, training-environment EM reaches $74.0\%$, while $S_0$ is only $24.0\%$.
Later in training, the attenuated feedback-assisted term declines while both $S_0$ and deployment EM rise.
By step 470, deployment probe EM reaches $51.0\%$, $S_0$ reaches $50.5\%$, average Privileged Feedback calls fall to $0.045$ per rollout, and only $4.0\%$ of probe rollouts invoke Privileged Feedback.
This pattern suggests that feedback-assisted repairs are being converted into behavior executable without Privileged Feedback calls.

\subsection{Attenuating Feedback Credit Improves Deployment Transfer}
\label{sec:exp_rho}

\begin{figure}[t]
\centering
\includegraphics[width=\linewidth]{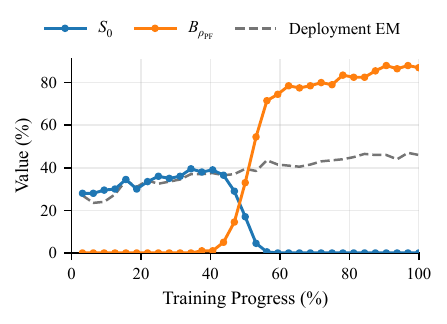}
\caption{Mechanism probe for unattenuated CAPF ($\rhopf=1.0$) on the same fixed 200-example set.
Training-environment success is decomposed into zero-feedback success $S_0$ and feedback-assisted contribution $B_{\rhopf}$ from Eq.~\eqref{eq:objective_decomp}.}
\label{fig:source_of_gain_unattenuated}
\vspace{-1.0ex}
\end{figure}

\begin{figure*}[t]
\centering
\includegraphics[width=\textwidth]{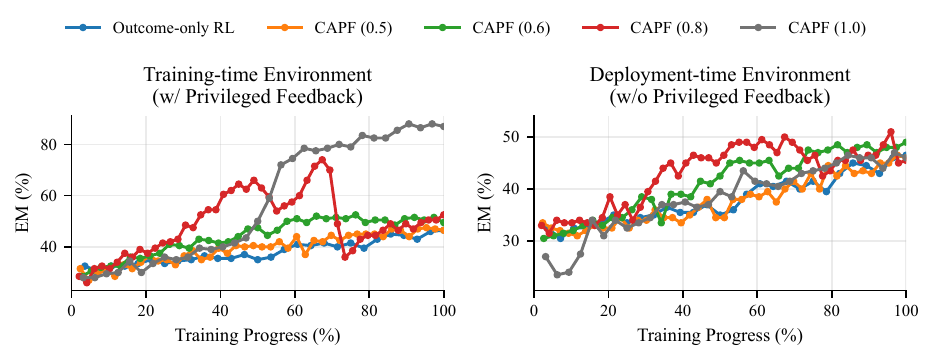}
\caption{Probe EM at matched checkpoints in the training and deployment environments.
Left: training environment, where the Privileged Feedback call is available to CAPF variants.
Right: deployment environment, where the Privileged Feedback call is removed.
The unattenuated $\rhopf=1.0$ setting reaches high training-environment EM but not the best deployment EM.}
\label{fig:training_dynamics}
\vspace{-1.0ex}
\end{figure*}

The comparison between $\rhopf=1.0$ and attenuated CAPF variants isolates the effect of credit attenuation.
All variants use the same Privileged Feedback call during training and remove it at evaluation.
With $\rhopf=1.0$, CAPF treats repaired trajectories as ordinary successes.
This setting improves average EM from $44.7\%$ to $46.1\%$ over outcome-only RL, showing that feedback-assisted trajectories provide a useful learning signal.
However, it underperforms all tested attenuated settings and trails $\rhopf=0.8$ by $2.4$ percentage points.

Figure~\ref{fig:training_dynamics} gives the corresponding probe view.
The unattenuated policy reaches high training-environment probe EM by continuing to call Privileged Feedback, but its deployment probe EM remains below the $\rhopf=0.8$ policy after this call is removed.
Figure~\ref{fig:source_of_gain_unattenuated} applies the same fixed-probe decomposition as Figure~\ref{fig:source_of_gain} to this unattenuated run.
Therefore, increasing training-environment success is not sufficient.

This gap reflects the credit problem CAPF is designed to address.
Post-feedback actions may contain transferable repair behavior, such as refined queries, better evidence selection, or repaired answers.
The feedback call and earlier actions should not receive the same credit as unassisted successful behavior because Privileged Feedback is absent at deployment.
CAPF therefore attenuates return whenever credit crosses a Privileged Feedback call, preserving learning signal from repair trajectories while reducing feedback-dependent credit.

\subsection{Probe Accounting for Tool Use and Search Budget}
\label{sec:exp_sanity}

\begin{table}[t]
\centering
\caption{Fixed-probe accounting at selected checkpoints.
Search and PF denote average calls per rollout.
Use~\% is the fraction of training-environment rollouts with at least one Privileged Feedback call.
Deployment metrics exclude Privileged Feedback.}
\label{tab:probe_accounting_main}
\smallskip
\begingroup
\scriptsize
\setlength{\tabcolsep}{2.6pt}
\resizebox{\linewidth}{!}{
\begin{tabular}{lrrrrrr}
\toprule
\multirow{2}{*}{Method}
& \multicolumn{4}{c}{\textbf{Train Env.}}
& \multicolumn{2}{c}{\textbf{Deploy Env.}} \\
\cmidrule(lr){2-5} \cmidrule(lr){6-7}
& EM (\%) & Search & PF & Use (\%)
& EM (\%) & Search \\
\midrule
Outcome-only RL
  & 46.5 & 2.45 & 0.000 & 0.0
  & 46.5 & 2.45 \\
CAPF ($\rhopf=0.5$)
  & 47.0 & 3.10 & 0.000 & 0.0
  & 46.5 & 2.17 \\
CAPF ($\rhopf=0.6$)
  & 49.5 & 10.41 & 0.130 & 3.5
  & 49.0 & 4.18 \\
CAPF ($\rhopf=0.8$)
  & 50.5 & 2.87 & 0.045 & 4.0
  & \textbf{51.0} & 1.98 \\
CAPF ($\rhopf=1.0$)
  & \textbf{88.0} & 7.51 & 6.455 & 100.0
  & 47.0 & 5.10 \\
\bottomrule
\end{tabular}}
\endgroup
\vspace{-1.0ex}
\end{table}

\begin{figure*}[!t]
\centering
\includegraphics[width=\textwidth]{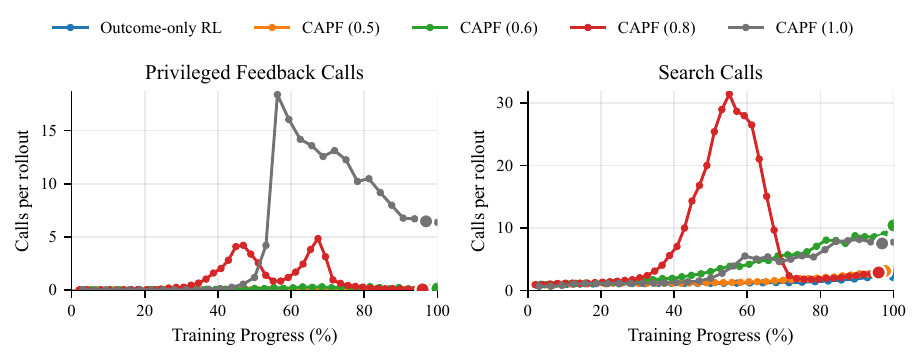}
\caption{Training environment overhead on the fixed probe.
Left: Privileged Feedback calls per rollout.
Right: search calls per rollout.
Enlarged markers indicate the checkpoints summarized in Table~\ref{tab:probe_accounting_main}.}
\label{fig:feedback_overhead}
\vspace{-1.0ex}
\end{figure*}

\begin{table*}[!t]
\centering
\caption{Comparison with recent search-agent RL systems on the seven-benchmark open-domain QA suite.
All scores are exact-match (EM) values in \%; Avg. denotes the benchmark-level macro-average.}
\label{tab:benchmark_positioning}
\smallskip
\begingroup
\small
\setlength{\tabcolsep}{2.8pt}
\begin{adjustbox}{max width=\textwidth}
\begin{tabular}{llcccccccc}
\toprule
\multirow{2}{*}{Method}
& \multirow{2}{*}{Base model}
& \multicolumn{3}{c}{\textbf{Single-hop QA}}
& \multicolumn{4}{c}{\textbf{Multi-hop QA}}
& \multirow{2}{*}{\textbf{Avg.}} \\
\cmidrule(lr){3-5} \cmidrule(lr){6-9}
& & NQ & TriviaQA & PopQA
& HotpotQA & 2Wiki & MuSiQue & Bamboogle & \\
\midrule
Search-o1-7B~\cite{li2025searcho1}
  & Qwen2.5-7B-Instruct
  & 15.1 & 44.3 & 13.1
  & 18.7 & 17.6 & 5.8 & 29.6 & 20.6 \\
Search-R1-7B~\cite{jin2025search}
  & Qwen2.5-7B-Instruct
  & 39.3 & 61.0 & 39.7
  & 37.0 & 41.4 & 14.6 & 36.8 & 38.5 \\
ZeroSearch-7B~\cite{sun2025zerosearch}
  & Qwen2.5-7B-Instruct
  & 43.6 & 65.2 & 48.8
  & 34.6 & 35.2 & 18.4 & 27.8 & 39.1 \\
TIPS-7B~\cite{xie2026tips}
  & Qwen2.5-7B-Instruct
  & 43.4 & 64.3 & 44.5
  & 43.0 & 43.0 & 17.1 & 36.8 & 41.7 \\
ReSeek-7B~\cite{li2025reseek}
  & Qwen2.5-7B-Instruct
  & 46.9 & 64.0 & 50.1
  & 38.9 & 38.2 & 18.5 & 39.2 & 42.3 \\
\midrule
CAPF-7B ($\rhopf=0.8$)
  & Qwen2.5-7B-Instruct
  & 44.8 & 63.6 & 43.6
  & 41.7 & 43.6 & 19.5 & 44.0 & 43.0 \\
\bottomrule
\end{tabular}
\end{adjustbox}
\endgroup
\vspace{-1.0ex}
\end{table*}

Table~\ref{tab:probe_accounting_main} reports checkpoint-level tool-use and search-budget accounting on the same 200-example probe.
The accounting checks whether the best deployment-probe result can be explained by larger search budgets or residual Privileged Feedback use.
CAPF with $\rhopf=0.8$ reaches $51.0\%$ deployment probe EM with $1.98$ search calls per rollout, compared with $46.5\%$ EM and $2.45$ search calls for outcome-only RL on the same probe.
Its training-environment Privileged Feedback use is also low, averaging $0.045$ calls per rollout, with only $4.0\%$ of rollouts invoking Privileged Feedback.
The unattenuated $\rhopf=1.0$ row provides a contrast point.
It reaches $88.0\%$ training-environment EM with a Privileged Feedback call in every probe rollout, but drops to $47.0\%$ deployment-probe EM after these calls are removed.
Thus, the strongest deployment-probe result is not explained by spending more search calls or by retaining the ability to call Privileged Feedback.

Figure~\ref{fig:feedback_overhead} shows the corresponding full fixed-probe trajectories for Privileged Feedback calls and search calls.
It complements Table~\ref{tab:probe_accounting_main}, which reports selected checkpoints only.
The curves distinguish two forms of overhead.
Privileged Feedback calls measure reliance on the call removed at deployment.
Search calls measure whether deployment gains coincide with larger retrieval budgets.
The unattenuated $\rho_{\mathrm{PF}}=1.0$ run shows sustained Privileged Feedback use.
The $\rho_{\mathrm{PF}}=0.6$ run shows higher search use at the selected checkpoint.
The $\rho_{\mathrm{PF}}=0.8$ run ends with low Privileged Feedback use and low deployment search use.

\subsection{Comparison with Recent Search-Agent RL Systems}
\label{sec:benchmark_comparison}

Table~\ref{tab:benchmark_positioning} compares CAPF with recent search-agent RL systems on the same seven-benchmark open-domain QA suite.
CAPF-7B uses Qwen2.5-7B-Instruct and follows the common Search-R1-style setup.
After removing Privileged Feedback, CAPF-7B achieves $43.0\%$ average EM, the highest macro-average among the listed Qwen2.5-7B systems.
It also obtains the strongest multi-hop results on 2Wiki, MuSiQue, and Bamboogle.
Several baselines remain stronger on some single-hop tasks and HotpotQA, so CAPF does not dominate every dataset.
Overall, CAPF remains competitive under deployment constraints.

\section{Conclusion}
\label{sec:conclusion}

We develop \textbf{Credit-Attenuated Privileged Feedback} (CAPF), a training-time mechanism for search-agent RLVR with verifier-side feedback. It adds a Privileged Feedback call during training and attenuates credit assigned to the feedback call and earlier actions, allowing repaired successes to provide a useful but attenuated learning signal.
Extensive experiments demonstrate that CAPF improves Qwen3-4B's average exact-match score from $44.7\%$ under outcome-only RLVR to $48.5\%$ across seven open-domain QA benchmarks without Privileged Feedback.
The results indicate that training-time verifier feedback can improve test-time search agents when its credit is properly controlled.

\section*{Limitations}
\label{sec:limitations}

Our experiments are limited to search-augmented open-domain QA with exact-match evaluation.
This setting provides automatic verifiers and relatively compact reference answers, making it suitable for studying Privileged Feedback.
We plan to explore CAPF's performance on tasks with longer outputs or higher subjective correctness in the future, as such tasks require reliable verifiers, feedback generators, and their tool schemas.

Another limitation is the potential leakage of Privileged Feedback.
Since Privileged Feedback is generated with access to the hidden reference answer, it may reveal answer-specific information even when it does not copy the reference string directly.
A more systematic measurement of such leakage is left for future work.

\bibliography{ref}

@inproceedings{chen2025research,
  title     = {{ReSearch}: Learning to Reason with Search for {LLMs} via Reinforcement Learning},
  author    = {Chen, Mingyang and Sun, Linzhuang and Li, Tianpeng and Sun, Haoze and Zhou, Yijie and Zhu, Chenzheng and Wang, Haofen and Pan, Jeff Z. and Zhang, Wen and Chen, Huajun and Yang, Fan and Zhou, Zenan and Chen, Weipeng},
  booktitle = {The Thirty-ninth Annual Conference on Neural Information Processing Systems},
  year      = {2025},
  url       = {https://openreview.net/forum?id=OuGAwwAT8G}
}

@inproceedings{choudhury2025leap,
  title         = {Better than Your Teacher: {LLM} Agents that learn from Privileged {AI} Feedback},
  author        = {Choudhury, Sanjiban and Sodhi, Paloma},
  booktitle     = {The Thirteenth International Conference on Learning Representations},
  year          = {2025},
  url           = {https://openreview.net/forum?id=st7XqFgbAH}
}

@misc{douze2024faiss,
  title         = {The {Faiss} Library},
  author        = {Douze, Matthijs and Guzhva, Alexandr and Deng, Chengqi and Johnson, Jeff and Szilvasy, Gergely and Mazar{\'e}, Pierre-Emmanuel and Lomeli, Maria and Hosseini, Lucas and J{\'e}gou, Herv{\'e}},
  year          = {2024},
  eprint        = {2401.08281},
  archivePrefix = {arXiv},
  doi           = {10.48550/arXiv.2401.08281},
  url           = {https://arxiv.org/abs/2401.08281}
}

@misc{gao2025beyond,
  title         = {Beyond Ten Turns: Unlocking Long-Horizon Agentic Search with Large-Scale Asynchronous {RL}},
  author        = {Gao, Jiaxuan and Fu, Wei and Xie, Minyang and Xu, Shusheng and He, Chuyi and Mei, Zhiyu and Zhu, Banghua and Wu, Yi},
  year          = {2025},
  eprint        = {2508.07976},
  archivePrefix = {arXiv},
  doi           = {10.48550/arXiv.2508.07976},
  url           = {https://arxiv.org/abs/2508.07976}
}

@article{guo2025deepseek,
  title   = {{DeepSeek-R1} incentivizes reasoning in {LLMs} through reinforcement learning},
  author  = {Guo, Daya and
             Yang, Dejian and
             Zhang, Haowei and
             Song, Junxiao and
             Wang, Peiyi and
             Zhu, Qihao and
             Xu, Runxin and
             Zhang, Ruoyu and
             Ma, Shirong and
             Bi, Xiao and
             Zhang, Xiaokang and
             Yu, Xingkai and
             Wu, Yu and
             Wu, Z. F. and
             Gou, Zhibin and
             Shao, Zhihong and
             Li, Zhuoshu and
             Gao, Ziyi and
             Liu, Aixin and
             Xue, Bing and
             Wang, Bingxuan and
             Wu, Bochao and
             Feng, Bei and
             Lu, Chengda and
             Zhao, Chenggang and
             Deng, Chengqi and
             Ruan, Chong and
             Dai, Damai and
             Chen, Deli and
             Ji, Dongjie and
             Li, Erhang and
             Lin, Fangyun and
             Dai, Fucong and
             Luo, Fuli and
             Hao, Guangbo and
             Chen, Guanting and
             Li, Guowei and
             Zhang, Hao and
             Xu, Hanwei and
             Ding, Honghui and
             Gao, Huazuo and
             Qu, Hui and
             Li, Hui and
             Guo, Jianzhong and
             Li, Jiashi and
             Chen, Jingchang and
             Yuan, Jingyang and
             Tu, Jinhao and
             Qiu, Junjie and
             Li, Junlong and
             Cai, J. L. and
             Ni, Jiaqi and
             Liang, Jian and
             Chen, Jin and
             Dong, Kai and
             Hu, Kai and
             You, Kaichao and
             Gao, Kaige and
             Guan, Kang and
             Huang, Kexin and
             Yu, Kuai and
             Wang, Lean and
             Zhang, Lecong and
             Zhao, Liang and
             Wang, Litong and
             Zhang, Liyue and
             Xu, Lei and
             Xia, Leyi and
             Zhang, Mingchuan and
             Zhang, Minghua and
             Tang, Minghui and
             Zhou, Mingxu and
             Li, Meng and
             Wang, Miaojun and
             Li, Mingming and
             Tian, Ning and
             Huang, Panpan and
             Zhang, Peng and
             Wang, Qiancheng and
             Chen, Qinyu and
             Du, Qiushi and
             Ge, Ruiqi and
             Zhang, Ruisong and
             Pan, Ruizhe and
             Wang, Runji and
             Chen, R. J. and
             Jin, R. L. and
             Chen, Ruyi and
             Lu, Shanghao and
             Zhou, Shangyan and
             Chen, Shanhuang and
             Ye, Shengfeng and
             Wang, Shiyu and
             Yu, Shuiping and
             Zhou, Shunfeng and
             Pan, Shuting and
             Li, S. S. and
             Zhou, Shuang and
             Wu, Shaoqing and
             Yun, Tao and
             Pei, Tian and
             Sun, Tianyu and
             Wang, Tao and
             Zeng, Wangding and
             Liu, Wen and
             Liang, Wenfeng and
             Gao, Wenjun and
             Yu, Wenqin and
             Zhang, Wentao and
             Xiao, W. L. and
             An, Wei and
             Liu, Xiaodong and
             Wang, Xiaohan and
             Chen, Xiaokang and
             Nie, Xiaotao and
             Cheng, Xin and
             Liu, Xin and
             Xie, Xin and
             Liu, Xingchao and
             Yang, Xinyu and
             Li, Xinyuan and
             Su, Xuecheng and
             Lin, Xuheng and
             Li, X. Q. and
             Jin, Xiangyue and
             Shen, Xiaojin and
             Chen, Xiaosha and
             Sun, Xiaowen and
             Wang, Xiaoxiang and
             Song, Xinnan and
             Zhou, Xinyi and
             Wang, Xianzu and
             Shan, Xinxia and
             Li, Y. K. and
             Wang, Y. Q. and
             Wei, Y. X. and
             Zhang, Yang and
             Xu, Yanhong and
             Li, Yao and
             Zhao, Yao and
             Sun, Yaofeng and
             Wang, Yaohui and
             Yu, Yi and
             Zhang, Yichao and
             Shi, Yifan and
             Xiong, Yiliang and
             He, Ying and
             Piao, Yishi and
             Wang, Yisong and
             Tan, Yixuan and
             Ma, Yiyang and
             Liu, Yiyuan and
             Guo, Yongqiang and
             Ou, Yuan and
             Wang, Yuduan and
             Gong, Yue and
             Zou, Yuheng and
             He, Yujia and
             Xiong, Yunfan and
             Luo, Yuxiang and
             You, Yuxiang and
             Liu, Yuxuan and
             Zhou, Yuyang and
             Zhu, Y. X. and
             Huang, Yanping and
             Li, Yaohui and
             Zheng, Yi and
             Zhu, Yuchen and
             Ma, Yunxian and
             Tang, Ying and
             Zha, Yukun and
             Yan, Yuting and
             Ren, Z. Z. and
             Ren, Zehui and
             Sha, Zhangli and
             Fu, Zhe and
             Xu, Zhean and
             Xie, Zhenda and
             Zhang, Zhengyan and
             Hao, Zhewen and
             Ma, Zhicheng and
             Yan, Zhigang and
             Wu, Zhiyu and
             Gu, Zihui and
             Zhu, Zijia and
             Liu, Zijun and
             Li, Zilin and
             Xie, Ziwei and
             Song, Ziyang and
             Pan, Zizheng and
             Huang, Zhen and
             Xu, Zhipeng and
             Zhang, Zhongyu and
             Zhang, Zhen},
  journal = {Nature},
  volume  = {645},
  number  = {8081},
  pages   = {633--638},
  year    = {2025},
  month   = sep,
  doi     = {10.1038/s41586-025-09422-z},
  url     = {https://doi.org/10.1038/s41586-025-09422-z}
}

@inproceedings{ho2020twikimultihopqa,
  title     = {Constructing A Multi-hop {QA} Dataset for Comprehensive Evaluation of Reasoning Steps},
  author    = {Ho, Xanh and Duong Nguyen, Anh-Khoa and Sugawara, Saku and Aizawa, Akiko},
  booktitle = {Proceedings of the 28th International Conference on Computational Linguistics},
  pages     = {6609--6625},
  year      = {2020},
  address   = {Barcelona, Spain (Online)},
  publisher = {International Committee on Computational Linguistics},
  doi       = {10.18653/v1/2020.coling-main.580},
  url       = {https://aclanthology.org/2020.coling-main.580/}
}

@inproceedings{hu2024openrlhf,
  title     = {{OpenRLHF}: A Ray-based Easy-to-use, Scalable and High-performance {RLHF} Framework},
  author    = {Hu, Jian and Wu, Xibin and Shen, Wei and Liu, Jason Klein and Wang, Weixun and Jiang, Songlin and Wang, Haoran and Chen, Hao and Chen, Bin and Fang, Wenkai and Xianyu and Cao, Yu and Xu, Haotian and Liu, Yiming},
  booktitle = {Proceedings of the 2025 Conference on Empirical Methods in Natural Language Processing: System Demonstrations},
  pages     = {656--666},
  year      = {2025},
  address   = {Suzhou, China},
  publisher = {Association for Computational Linguistics},
  doi       = {10.18653/v1/2025.emnlp-demos.48},
  url       = {https://aclanthology.org/2025.emnlp-demos.48/}
}

@misc{hu2025reinforcepp,
  title         = {{REINFORCE++}: Stabilizing Critic-Free Policy Optimization with Global Advantage Normalization},
  author        = {Hu, Jian and Liu, Jason Klein and Xu, Haotian and Shen, Wei},
  year          = {2025},
  eprint        = {2501.03262},
  archivePrefix = {arXiv},
  doi           = {10.48550/arXiv.2501.03262},
  url           = {https://arxiv.org/abs/2501.03262}
}

@inproceedings{jin2024flashrag,
  title     = {{FlashRAG}: A Modular Toolkit for Efficient Retrieval-Augmented Generation Research},
  author    = {Jin, Jiajie and Zhu, Yutao and Dou, Zhicheng and Dong, Guanting and Yang, Xinyu and Zhang, Chenghao and Zhao, Tong and Yang, Zhao and Wen, Ji-Rong},
  booktitle = {Companion Proceedings of the {ACM} on Web Conference 2025},
  pages     = {737--740},
  year      = {2025},
  month     = may,
  publisher = {ACM},
  doi       = {10.1145/3701716.3715313},
  url       = {https://doi.org/10.1145/3701716.3715313}
}

@inproceedings{jin2025search,
  title     = {{Search-R1}: Training {LLMs} to Reason and Leverage Search Engines with Reinforcement Learning},
  author    = {Jin, Bowen and Zeng, Hansi and Yue, Zhenrui and Yoon, Jinsung and Arik, Sercan O. and Wang, Dong and Zamani, Hamed and Han, Jiawei},
  booktitle = {Proceedings of the Second Conference on Language Modeling},
  year      = {2025},
  url       = {https://openreview.net/forum?id=Rwhi91ideu}
}

@inproceedings{joshi2017triviaqa,
  title     = {{TriviaQA}: A Large Scale Distantly Supervised Challenge Dataset for Reading Comprehension},
  author    = {Joshi, Mandar and Choi, Eunsol and Weld, Daniel and Zettlemoyer, Luke},
  booktitle = {Proceedings of the 55th Annual Meeting of the Association for Computational Linguistics (Volume 1: Long Papers)},
  pages     = {1601--1611},
  year      = {2017},
  address   = {Vancouver, Canada},
  publisher = {Association for Computational Linguistics},
  doi       = {10.18653/v1/P17-1147},
  url       = {https://aclanthology.org/P17-1147/}
}

@article{kwiatkowski2019natural,
  title     = {Natural Questions: A Benchmark for Question Answering Research},
  author    = {Kwiatkowski, Tom and Palomaki, Jennimaria and Redfield, Olivia and Collins, Michael and Parikh, Ankur and Alberti, Chris and Epstein, Danielle and Polosukhin, Illia and Devlin, Jacob and Lee, Kenton and Toutanova, Kristina and Jones, Llion and Kelcey, Matthew and Chang, Ming-Wei and Dai, Andrew M. and Uszkoreit, Jakob and Le, Quoc and Petrov, Slav},
  journal   = {Transactions of the Association for Computational Linguistics},
  volume    = {7},
  pages     = {452--466},
  year      = {2019},
  address   = {Cambridge, MA},
  publisher = {MIT Press},
  doi       = {10.1162/tacl_a_00276},
  url       = {https://aclanthology.org/Q19-1026/}
}

@misc{li2025reseek,
  title         = {{ReSeek}: A Self-Correcting Framework for Search Agents with Instructive Rewards},
  author        = {Li, Shiyu and Tang, Yang and Wang, Yifan and Li, Peiming and Chen, Xi},
  year          = {2025},
  eprint        = {2510.00568},
  archivePrefix = {arXiv},
  doi           = {10.48550/arXiv.2510.00568},
  url           = {https://arxiv.org/abs/2510.00568}
}

@inproceedings{li2025searcho1,
  title     = {{Search-o1}: Agentic Search-Enhanced Large Reasoning Models},
  author    = {Li, Xiaoxi and Dong, Guanting and Jin, Jiajie and Zhang, Yuyao and Zhou, Yujia and Zhu, Yutao and Zhang, Peitian and Dou, Zhicheng},
  booktitle = {Proceedings of the 2025 Conference on Empirical Methods in Natural Language Processing},
  pages     = {5420--5438},
  year      = {2025},
  address   = {Suzhou, China},
  publisher = {Association for Computational Linguistics},
  doi       = {10.18653/v1/2025.emnlp-main.276},
  url       = {https://aclanthology.org/2025.emnlp-main.276/}
}

@inproceedings{lightman2023let,
  title         = {Let's Verify Step by Step},
  author        = {Lightman, Hunter and Kosaraju, Vineet and Burda, Yuri and Edwards, Harrison and Baker, Bowen and Lee, Teddy and Leike, Jan and Schulman, John and Sutskever, Ilya and Cobbe, Karl},
  booktitle     = {The Twelfth International Conference on Learning Representations},
  year          = {2024},
  url           = {https://openreview.net/forum?id=v8L0pN6EOi}
}

@inproceedings{lu2026searchselfplay,
  title         = {Search Self-Play: Pushing the Frontier of Agent Capability without Supervision},
  author        = {Lu, Hongliang and Wen, Yuhang and Cheng, Pengyu and Ding, Ruijin and Guo, Jiaqi and Xu, Haotian and Wang, Chutian and Chen, Haonan and Jiang, Xiaoxi and Jiang, Guanjun},
  booktitle     = {The Fourteenth International Conference on Learning Representations},
  year          = {2026},
  url           = {https://openreview.net/forum?id=ZmGirmNJqE}
}

@misc{luo2025infoflow,
  title         = {{InfoFlow}: Reinforcing Search Agent Via Reward Density Optimization},
  author        = {Luo, Kun and Qian, Hongjin and Liu, Zheng and Xia, Ziyi and Xiao, Shitao and Bao, Siqi and Zhao, Jun and Liu, Kang},
  year          = {2025},
  eprint        = {2510.26575},
  archivePrefix = {arXiv},
  doi           = {10.48550/arXiv.2510.26575},
  url           = {https://arxiv.org/abs/2510.26575}
}

@inproceedings{mallen2022whennottrust,
  title     = {When Not to Trust Language Models: Investigating Effectiveness of Parametric and Non-Parametric Memories},
  author    = {Mallen, Alex and Asai, Akari and Zhong, Victor and Das, Rajarshi and Khashabi, Daniel and Hajishirzi, Hannaneh},
  booktitle = {Proceedings of the 61st Annual Meeting of the Association for Computational Linguistics (Volume 1: Long Papers)},
  pages     = {9802--9822},
  year      = {2023},
  month     = jul,
  address   = {Toronto, Canada},
  publisher = {Association for Computational Linguistics},
  doi       = {10.18653/v1/2023.acl-long.546},
  url       = {https://aclanthology.org/2023.acl-long.546/}
}

@misc{nakano2021webgpt,
  title         = {{WebGPT}: Browser-assisted question-answering with human feedback},
  author        = {Nakano, Reiichiro and Hilton, Jacob and Balaji, Suchir and Wu, Jeff and Ouyang, Long and Kim, Christina and Hesse, Christopher and Jain, Shantanu and Kosaraju, Vineet and Saunders, William and Jiang, Xu and Cobbe, Karl and Eloundou, Tyna and Krueger, Gretchen and Button, Kevin and Knight, Matthew and Chess, Benjamin and Schulman, John},
  year          = {2021},
  eprint        = {2112.09332},
  archivePrefix = {arXiv},
  doi           = {10.48550/arXiv.2112.09332},
  url           = {https://arxiv.org/abs/2112.09332}
}

@inproceedings{ng1999policy,
  title     = {Policy Invariance under Reward Transformations: Theory and Application to Reward Shaping},
  author    = {Ng, Andrew Y. and Harada, Daishi and Russell, Stuart},
  booktitle = {Proceedings of the Sixteenth International Conference on Machine Learning},
  pages     = {278--287},
  year      = {1999},
  url       = {https://people.eecs.berkeley.edu/~pabbeel/cs287-fa09/readings/NgHaradaRussell-shaping-ICML1999.pdf}
}

@inproceedings{sutton1999policy,
  title     = {Policy Gradient Methods for Reinforcement Learning with Function Approximation},
  author    = {Sutton, Richard S. and McAllester, David and Singh, Satinder and Mansour, Yishay},
  booktitle = {Advances in Neural Information Processing Systems},
  volume    = {12},
  pages     = {1057--1063},
  year      = {1999},
  publisher = {MIT Press},
  url       = {https://proceedings.neurips.cc/paper_files/paper/1999/hash/464d828b85b0bed98e80ade0a5c43b0f-Abstract.html}
}

@inproceedings{pechyony2010lupi,
  title     = {On the Theory of Learning with Privileged Information},
  author    = {Pechyony, Dmitry and Vapnik, Vladimir},
  booktitle = {Advances in Neural Information Processing Systems},
  volume    = {23},
  pages     = {1894--1902},
  year      = {2010},
  publisher = {Curran Associates, Inc.},
  url       = {https://papers.neurips.cc/paper/3960-on-the-theory-of-learnining-with-privileged-information}
}

@inproceedings{pitis2019rethinking,
  title     = {Rethinking the Discount Factor in Reinforcement Learning: A Decision Theoretic Approach},
  author    = {Pitis, Silviu},
  booktitle = {Proceedings of the AAAI Conference on Artificial Intelligence},
  volume    = {33},
  pages     = {7949--7956},
  year      = {2019},
  doi       = {10.1609/aaai.v33i01.33017949},
  url       = {https://ojs.aaai.org/index.php/AAAI/article/view/4795}
}

@inproceedings{press2022compositionalitygap,
  title     = {Measuring and Narrowing the Compositionality Gap in Language Models},
  author    = {Press, Ofir and Zhang, Muru and Min, Sewon and Schmidt, Ludwig and Smith, Noah and Lewis, Mike},
  booktitle = {Findings of the Association for Computational Linguistics: EMNLP 2023},
  pages     = {5687--5711},
  year      = {2023},
  address   = {Singapore},
  publisher = {Association for Computational Linguistics},
  doi       = {10.18653/v1/2023.findings-emnlp.378},
  url       = {https://aclanthology.org/2023.findings-emnlp.378/}
}

@misc{qwen2024qwen25,
  title         = {{Qwen2.5} Technical Report},
  author        = {{Qwen Team}},
  year          = {2024},
  eprint        = {2412.15115},
  archivePrefix = {arXiv},
  doi           = {10.48550/arXiv.2412.15115},
  url           = {https://arxiv.org/abs/2412.15115}
}

@misc{qwen2025qwen3,
  title         = {{Qwen3} Technical Report},
  author        = {{Qwen Team}},
  year          = {2025},
  eprint        = {2505.09388},
  archivePrefix = {arXiv},
  doi           = {10.48550/arXiv.2505.09388},
  url           = {https://arxiv.org/abs/2505.09388}
}

@inproceedings{qu2026pope,
  title         = {Learning to Reason on Hard Problems with Privileged On-Policy Exploration},
  author        = {Qu, Yuxiao and Setlur, Amrith and Smith, Virginia and Salakhutdinov, Ruslan and Kumar, Aviral},
  booktitle     = {The 5th Workshop on Mathematical Reasoning and AI at NeurIPS 2025},
  year          = {2025},
  url           = {https://openreview.net/forum?id=zKn6mVwPZE}
}

@inproceedings{shi2025autorefine,
  title         = {Search and Refine During Think: Facilitating Knowledge Refinement for Improved Retrieval-Augmented Reasoning},
  author        = {Shi, Yaorui and Li, Sihang and Wu, Chang and Liu, Zhiyuan and Fang, Junfeng and Cai, Hengxing and Zhang, An and Wang, Xiang},
  booktitle     = {Advances in Neural Information Processing Systems},
  volume        = {38},
  year          = {2025},
  url           = {https://papers.nips.cc/paper_files/paper/2025/hash/e4a7f5019f06d997e7e8936fcc948c85-Abstract-Conference.html}
}

@misc{sun2025zerosearch,
  title         = {{ZeroSearch}: Incentivize the Search Capability of {LLMs} without Searching},
  author        = {Sun, Hao and Qiao, Zile and Guo, Jiayan and Fan, Xuanbo and Hou, Yingyan and Jiang, Yong and Xie, Pengjun and Zhang, Yan and Huang, Fei and Zhou, Jingren},
  year          = {2025},
  eprint        = {2505.04588},
  archivePrefix = {arXiv},
  doi           = {10.48550/arXiv.2505.04588},
  url           = {https://arxiv.org/abs/2505.04588}
}

@article{trivedi2021musique,
  title     = {{M}u{S}i{Q}ue: Multihop Questions via Single-hop Question Composition},
  author    = {Trivedi, Harsh and Balasubramanian, Niranjan and Khot, Tushar and Sabharwal, Ashish},
  journal   = {Transactions of the Association for Computational Linguistics},
  volume    = {10},
  pages     = {539--554},
  year      = {2022},
  address   = {Cambridge, MA},
  publisher = {MIT Press},
  doi       = {10.1162/tacl_a_00475},
  url       = {https://aclanthology.org/2022.tacl-1.31/}
}

@misc{uesato2022process,
  title         = {Solving math word problems with process- and outcome-based feedback},
  author        = {Uesato, Jonathan and Kushman, Nate and Kumar, Ramana and Song, Francis and Siegel, Noah and Wang, Lisa and Creswell, Antonia and Irving, Geoffrey and Higgins, Irina},
  year          = {2022},
  eprint        = {2211.14275},
  archivePrefix = {arXiv},
  doi           = {10.48550/arXiv.2211.14275},
  url           = {https://arxiv.org/abs/2211.14275}
}

@article{wang2022text,
  title   = {Text Embeddings by Weakly-Supervised Contrastive Pre-training},
  author  = {Wang, Liang and Yang, Nan and Huang, Xiaolong and Jiao, Binxing and Yang, Linjun and Jiang, Daxin and Majumder, Rangan and Wei, Furu},
  journal = {arXiv preprint arXiv:2212.03533},
  year    = {2022},
  url     = {https://arxiv.org/abs/2212.03533}
}

@article{vapnik2009lupi,
  title   = {A New Learning Paradigm: Learning Using Privileged Information},
  author  = {Vapnik, Vladimir and Vashist, Akshay},
  journal = {Neural Networks},
  volume  = {22},
  number  = {5--6},
  pages   = {544--557},
  year    = {2009},
  month   = jul,
  doi     = {10.1016/j.neunet.2009.06.042},
  url     = {https://doi.org/10.1016/j.neunet.2009.06.042}
}

@misc{wen2025rlvr,
  title         = {Reinforcement Learning with Verifiable Rewards Implicitly Incentivizes Correct Reasoning in Base {LLMs}},
  author        = {Wen, Xumeng and Liu, Zihan and Zheng, Shun and Ye, Shengyu and Wu, Zhirong and Wang, Yang and Xu, Zhijian and Liang, Xiao and Li, Junjie and Miao, Ziming and Bian, Jiang and Yang, Mao},
  year          = {2025},
  eprint        = {2506.14245},
  archivePrefix = {arXiv},
  doi           = {10.48550/arXiv.2506.14245},
  url           = {https://arxiv.org/abs/2506.14245}
}

@inproceedings{white2017unifying,
  title     = {Unifying Task Specification in Reinforcement Learning},
  author    = {White, Martha},
  booktitle = {Proceedings of the 34th International Conference on Machine Learning},
  pages     = {3742--3750},
  year      = {2017},
  url       = {https://proceedings.mlr.press/v70/white17a.html}
}

@article{williams1992simple,
  title   = {Simple Statistical Gradient-Following Algorithms for Connectionist Reinforcement Learning},
  author  = {Williams, Ronald J.},
  journal = {Machine Learning},
  volume  = {8},
  number  = {3--4},
  pages   = {229--256},
  year    = {1992},
  month   = may,
  doi     = {10.1007/BF00992696},
  url     = {https://doi.org/10.1007/BF00992696}
}

@inproceedings{xie2026tips,
  title         = {{TIPS}: Turn-level Information-Potential Reward Shaping for Search-Augmented {LLMs}},
  author        = {Xie, Yutao and Thomas, Nathaniel and Hansen, Nicklas and Fu, Yang and Li, Li Erran and Wang, Xiaolong},
  booktitle     = {The Fourteenth International Conference on Learning Representations},
  year          = {2026},
  url           = {https://openreview.net/forum?id=eBMOr6a84z}
}

@inproceedings{yang2018hotpotqa,
  title     = {{HotpotQA}: A Dataset for Diverse, Explainable Multi-hop Question Answering},
  author    = {Yang, Zhilin and Qi, Peng and Zhang, Saizheng and Bengio, Yoshua and Cohen, William and Salakhutdinov, Ruslan and Manning, Christopher D.},
  booktitle = {Proceedings of the 2018 Conference on Empirical Methods in Natural Language Processing},
  pages     = {2369--2380},
  year      = {2018},
  address   = {Brussels, Belgium},
  publisher = {Association for Computational Linguistics},
  doi       = {10.18653/v1/D18-1259},
  url       = {https://aclanthology.org/D18-1259/}
}

@inproceedings{yao2023react,
  title         = {{ReAct}: Synergizing Reasoning and Acting in Language Models},
  author        = {Yao, Shunyu and Zhao, Jeffrey and Yu, Dian and Du, Nan and Shafran, Izhak and Narasimhan, Karthik R. and Cao, Yuan},
  booktitle     = {The Eleventh International Conference on Learning Representations},
  year          = {2023},
  url           = {https://openreview.net/forum?id=WE_vluYUL-X}
}

@inproceedings{zheng2025stepsearch,
  title     = {{StepSearch}: Igniting {LLMs} Search Ability via Step-Wise Proximal Policy Optimization},
  author    = {Zheng, Xuhui and An, Kang and Wang, Ziliang and Wang, Yuhang and Wu, Yichao},
  booktitle = {Proceedings of the 2025 Conference on Empirical Methods in Natural Language Processing},
  pages     = {21805--21830},
  year      = {2025},
  address   = {Suzhou, China},
  publisher = {Association for Computational Linguistics},
  doi       = {10.18653/v1/2025.emnlp-main.1106},
  url       = {https://aclanthology.org/2025.emnlp-main.1106/}
}

@inproceedings{zheng2025deepresearcher,
  title     = {{DeepResearcher}: Scaling Deep Research via Reinforcement Learning in Real-world Environments},
  author    = {Zheng, Yuxiang and Fu, Dayuan and Hu, Xiangkun and Cai, Xiaojie and Ye, Lyumanshan and Lu, Pengrui and Liu, Pengfei},
  booktitle = {Proceedings of the 2025 Conference on Empirical Methods in Natural Language Processing},
  pages     = {414--431},
  year      = {2025},
  address   = {Suzhou, China},
  publisher = {Association for Computational Linguistics},
  doi       = {10.18653/v1/2025.emnlp-main.22},
  url       = {https://aclanthology.org/2025.emnlp-main.22/}
}

\clearpage
\appendix

\section{Data and Benchmarks}
\label{app:data_eval}

\subsection{Training Data}
\label{app:training_data}

All runs use \texttt{PeterJinGo/nq\_hotpotqa\_train}, the same training set used by Search-R1~\cite{jin2025search}.
The dataset is formed from the NQ and HotpotQA training splits in FlashRAG~\cite{jin2024flashrag}.
It contains $169{,}615$ question--answer instances in total ($79{,}168$ from NQ and $90{,}447$ from HotpotQA).
We use the same data sampling script for outcome-only RL and all CAPF variants.
No examples from the evaluation splits in Table~\ref{tab:evaluation_benchmarks} are used for policy-gradient updates.

\subsection{Evaluation Benchmarks}
\label{app:evaluation_benchmarks}

The evaluation suite contains seven open-domain QA benchmarks.
Following recent search-agent RL work~\cite{jin2025search,li2025searcho1,xie2026tips,li2025reseek}, we group NQ, TriviaQA, and PopQA as single-hop QA, and HotpotQA, 2WikiMultiHopQA, MuSiQue, and Bamboogle as multi-hop QA.
Table~\ref{tab:evaluation_benchmarks} reports the source split names and example counts from the FlashRAG release used to construct this suite~\cite{jin2024flashrag}.
All reported CAPF benchmark numbers are measured after removing the Privileged Feedback call.

We report exact-match (EM) scores in \%.
All reported averages are benchmark-level macro-averages over EM values.
The single-hop and multi-hop averages use the same equal-benchmark convention within their respective groups.
For final-answer parsing and EM normalization, we follow Search-R1~\cite{jin2025search}.
Predictions are lowercased with articles and punctuation removed.
Multi-answer examples are counted as correct when any normalized reference matches the prediction.
Rollouts that do not produce a valid final-answer field before the turn budget is exhausted receive reward zero.
The same parser and normalizer are used for both training rewards and deployment evaluation.

\begin{table}[t]
\centering
\caption{Evaluation benchmark categories, splits, and example counts.}
\label{tab:evaluation_benchmarks}
\smallskip
\begingroup
\small
\setlength{\tabcolsep}{2.5pt}
\renewcommand{\arraystretch}{0.98}
\begin{tabular}{llll}
\toprule
\textbf{Benchmark} & \textbf{Category} & \textbf{Split} & \textbf{\# examples} \\
\midrule
NQ & Single-hop & test & $3{,}610$ \\
TriviaQA & Single-hop & test & $11{,}313$ \\
PopQA & Single-hop & test & $14{,}267$ \\
HotpotQA & Multi-hop & dev & $7{,}405$ \\
2WikiMultiHopQA & Multi-hop & dev & $12{,}576$ \\
MuSiQue & Multi-hop & dev & $2{,}417$ \\
Bamboogle & Multi-hop & test & $125$ \\
\bottomrule
\end{tabular}
\endgroup
\vspace{-1.0ex}
\end{table}

\section{Training Protocol}
\label{app:training_protocol}

This appendix documents the search environment, compute setup, and optimization settings used for the learned runs.
The settings apply to the controlled Qwen3-4B comparison in Table~\ref{tab:qa_matched}.
The Qwen2.5-7B run in Table~\ref{tab:benchmark_positioning} follows the same protocol unless otherwise noted.

Training was conducted on a three-node cluster with eight NVIDIA A800 GPUs per node.
One node was used for actor and reference-model training, and the remaining two nodes were used for rollout generation.
A full large-scale learned run takes approximately five days on this cluster, corresponding to about $2{,}880$ A800 GPU-hours per run.

\subsection{Retrieval Backend}
\label{app:retrieval_backend}

All methods use the same local Search-R1 retrieval configuration~\cite{jin2025search}.
The retrieval server uses a FAISS index~\cite{douze2024faiss} over the Wiki-18 corpus, with queries encoded by \texttt{intfloat/e5-base-v2}~\cite{wang2022text}.
Each \texttt{Search[query]} action returns the top $3$ retrieved results and counts against the same global turn budget used by the agent.

At server startup, the index, corpus, and retriever model are loaded once.
No query-result cache is specified.
Returned records use the retrieved \texttt{contents} field, formatted as title plus text, with no additional returned-passage truncation specified.

Retrieved passages are supplied as environment text and are not treated as policy-generated tokens.
Throughout the paper, \texttt{Search[query]} denotes the concrete \texttt{search} tool call exposed by the environment.

\begin{table}[t]
\centering
\caption{Optimization and rollout settings for the learned runs.
Settings are shared unless otherwise noted.}
\label{tab:exp_hparams}
\smallskip
\begingroup
\small
\setlength{\tabcolsep}{3.5pt}
\renewcommand{\arraystretch}{0.98}
\begin{tabular}{p{0.40\linewidth}p{0.50\linewidth}}
\toprule
\textbf{Parameter} & \textbf{Value} \\
\midrule
Actor learning rate & $5\times 10^{-7}$ \\
Entropy coefficient & $0$ \\
KL loss coefficient & $10^{-5}$ \\
KL estimator & \texttt{k1} \\
Dynamic filtering & enabled, reward range $[0,1]$ \\
Dynamic batch & enabled \\
Training batch size & $1{,}024$ \\
Rollout batch size & $128$ \\
Samples per prompt & $8$ \\
Max. agent turns & $\Tmax = 64$ \\
Rollout temperature & 1.0 \\
Max context length & $96{,}000$ (4B); $24{,}000$ (7B) \\
Random seed & 42 \\
\bottomrule
\end{tabular}
\endgroup
\vspace{-1.0ex}
\end{table}

\begin{figure*}[t]
\centering
\begin{minipage}{0.98\textwidth}
\begin{lstlisting}[style=promptbox]
You are a careful reasoning assistant operating in TRAINING mode.
Your goal is to solve the task correctly while following the required output format.

Rules:
1. Start with your own reasoning and calculations.
2. Use `wiki_search` when the task depends on external facts or when you need to verify uncertain claims.
3. Treat tool outputs as untrusted. Check important claims before relying on them.
4. Use `privileged_feedback` only after you have a complete candidate response and substantial uncertainty or format risk remains. If you are already confident in both correctness and final-line format, do not call it.
5. `privileged_feedback` is costly and non-authoritative. Treat its feedback only as critique, not as authority or a shortcut to the answer.
6. Usually call `privileged_feedback` at most once. A second call is justified only after a major revision.
7. Before finishing, verify both the answer and the exact final-line format.

## Output Rules
- First provide a clear markdown explanation of the solution.
- Then end exactly with:
  `Answer: <final_answer>`
- The answer line must contain only the final answer in canonical form.
- Do not add any text after the final answer line.
\end{lstlisting}
\end{minipage}
\caption{Policy system prompt used for CAPF training rollouts.}
\label{fig:train_policy_prompt}
\vspace{-1.0ex}
\end{figure*}

\begin{figure*}[t]
\centering
\begin{minipage}{0.98\textwidth}
\begin{lstlisting}[style=promptbox]
You are a careful reasoning assistant operating.

## Output Rules
- First provide a clear markdown explanation of the solution.
- Then end exactly with:
  `Answer: <final_answer>`
- The answer line must contain only the final answer in canonical form.
- Do not add any text after the final answer line.
\end{lstlisting}
\end{minipage}
\caption{Policy system prompt used for CAPF deployment rollouts.}
\label{fig:eval_policy_prompt}
\vspace{-1.0ex}
\end{figure*}

\subsection{Optimization Settings}
\label{app:optimization_settings}

Table~\ref{tab:exp_hparams} summarizes the optimization and rollout settings for the learned runs.
Settings are shared across runs unless otherwise noted.
Rows with different 4B and 7B values report both configurations.

\subsection{Dynamic Filtering Rule}
\label{app:dynamic_filtering}

Dynamic filtering is enabled for all learned rows.
For each prompt, multiple rollouts are sampled and scored by the final-answer exact-match verifier.
Outcome-only RL and CAPF use the same dynamic-filtering range.

The filtering rule is matched, but the realized retained rollouts can differ across runs.
Thus, the controlled comparison in Table~\ref{tab:qa_matched} matches the retrieval environment, rollout protocol, and filtering rule, but not the exact number of retained gradient-carrying samples.

\section{CAPF Implementation}
\label{app:capf_impl}

The main text defines the Privileged Feedback call and the CAPF return.
This appendix documents the implementation details, including prompts, the tool schema, redaction, and rollout traces.

\subsection{Policy System Prompts}
\label{app:policy_prompts}

Figures~\ref{fig:train_policy_prompt} and~\ref{fig:eval_policy_prompt} give the system prompts used for CAPF training rollouts and deployment evaluation.
The \texttt{wiki\_search} tool schema is supplied by the environment rather than by the system prompt, and is shared across outcome-only training, CAPF training, and deployment evaluation.
Outcome-only RL is trained with the deployment prompt.
CAPF training keeps the same answer-format contract but additionally exposes the \texttt{privileged\_feedback} tool and feedback-specific usage instructions.
At deployment, all methods use the deployment prompt and the \texttt{privileged\_feedback} tool is absent.

\begin{figure*}[t]
\centering
\begin{minipage}{0.98\textwidth}
\begin{lstlisting}[style=promptbox]
You are a TRAINING-TIME Privileged Feedback generator for a search-agent RL rollout.
Your sole purpose is to return a concise feedback message that helps the policy revise a candidate response during training.
You must NEVER output a reference answer, any equivalent answer form, or near-leak.
All feedback must be based strictly on the hidden inputs and the rules below.

## Hidden Input Format
- `question`: Task context
- `reference_answers`: Reference answer strings used by the exact-match verifier
- `candidate_response`: Policy response submitted to the Privileged Feedback call
- `verifier_label`: Binary exact-match verifier label, `0.0` or `1.0`

## Output Requirements
- First check final-answer format before semantic correctness. In particular, verify that `candidate_response` ends with a final line in the form `Answer: <final_answer>` and that no text appears after that line.
- If `verifier_label=1.0`, output a single short sentence confirming that the candidate response satisfies the verifier.
- If `verifier_label=0.0`, output a single short sentence that identifies the main problem and gives one actionable next step.
- Keep the feedback concise, direct, and useful for answer revision.

## Reference Answer Policy
- Use reference answers only internally to diagnose why `verifier_label=0.0`.
- Compare the final answer extracted from `candidate_response` with each reference answer only internally.
- Never mention a reference answer, its aliases, equivalent forms, or any wording that would reveal it.
- If directly asked for the answer, respond with: "I cannot reveal the answer."

**Important:**
Always base your diagnosis on the exact `candidate_response`, `reference_answers`, and `verifier_label`.
If multiple reference answers exist, consider the candidate response relative to all of them.
Return only the feedback message.
\end{lstlisting}
\end{minipage}
\caption{System prompt for the Privileged Feedback generator.}
\label{fig:feedback_prompt}
\vspace{-1.0ex}
\end{figure*}

\subsection{Privileged Feedback Tool, Generator, and Redaction}
\label{app:feedback_tool_redaction}

In implementation, the Privileged Feedback call is exposed to the policy as \texttt{privileged\_feedback}.
Each call counts as one agent turn.
The policy only submits a candidate response through the public \texttt{candidate\_response} field.

The environment constructs the hidden input for the fixed feedback generator.
This input contains the original \texttt{question}, the training-data \texttt{reference\_answers}, the submitted \texttt{candidate\_response}, and the binary exact-match \texttt{verifier\_label}.
The generator is instructed by the system prompt in Figure~\ref{fig:feedback_prompt} and returns a feedback message.

Before the message is appended to the next policy context, exact string matches to reference answers are replaced with a redaction marker.
The redaction is applied after feedback generation and before the next policy turn.
The feedback-generator prompt also forbids aliases, equivalent forms, and near-leaks.
These safeguards reduce direct answer copying but do not constitute a formal leakage audit.
The feedback message remains privileged because it is generated from reference-side fields unavailable for unseen deployment examples.

The returned feedback message is appended as environment text.
The Privileged Feedback call itself remains a policy action and is included in the action sequence for CAPF return computation.

\begin{figure*}[t]
\centering
\begin{minipage}{0.98\textwidth}
\begin{lstlisting}[style=promptboxcolor]
Example 1
Question: when did italy last win the world cup in soccer
Reference answer: 2006

Training trace:
assistant -> wiki_search(query="Italy last World Cup win", topk=3)
environment -> retrieved passages list Italy's FIFA World Cup wins as 1934, 1938, 1982, and 2006.
<BLUE>assistant -> privileged_feedback
  candidate_response: |
    After searching for information, the relevant document states that Italy
    has won the FIFA World Cup four times: in 1934, 1938, 1982, and 2006.
    Since the most recent World Cup win for Italy is 2006, the answer is 2006.
    Answer: 2006</BLUE>
<GREEN>environment -> feedback: "Great! Your answer is correct."</GREEN>
assistant -> Answer: 2006

Deployment trace:
assistant -> wiki_search(query="Italy last World Cup win year", topk=3)
environment -> retrieved passages list Italy's FIFA World Cup wins as 1934, 1938, 1982, and 2006.
assistant -> Answer: 2006
\end{lstlisting}
\end{minipage}
\caption{Abridged rollout trace for a correct candidate during CAPF training.
The training trace can call Privileged Feedback for confirmation, while the deployment trace runs without Privileged Feedback calls.}
\label{fig:rollout_trace_example}
\vspace{-1.0ex}
\end{figure*}

\begin{figure*}[t]
\centering
\begin{minipage}{0.98\textwidth}
\begin{lstlisting}[style=promptboxcolor]
Example 2
Question: the area in which crop is grown at a particular time or crop season is called?
Reference answer: growing season

Training trace:
assistant -> wiki_search(query="area in which crop is grown at a particular time or crop season", topk=3)
environment -> retrieved passages include definitions for growing season and growing region.
<BLUE>assistant -> privileged_feedback
  candidate_response: |
    The term for the area in which crops are grown at a particular time
    or crop season is known as a growing region.
    Answer: growing region</BLUE>
<GREEN>environment -> feedback: "The candidate response focuses on a geographical area rather than the time period asked for. Revise the final answer."</GREEN>
<BLUE>assistant -> privileged_feedback
  candidate_response: |
    The term for the time period during which crops are grown is known
    as a growing season.
    Answer: growing season</BLUE>
<GREEN>environment -> feedback: "Great! Your answer is correct."</GREEN>
assistant -> Answer: growing season

Deployment trace:
assistant -> wiki_search(query="agricultural term for field where crops are grown during a season", topk=5)
environment -> retrieved passages include field, growing season, and growing region.
<RED>assistant -> Answer: field</RED>
\end{lstlisting}
\end{minipage}
\caption{Abridged rollout trace for feedback-assisted correction during CAPF training.
The training trace repairs an initially wrong candidate after Privileged Feedback, while the deployment trace runs without Privileged Feedback calls and keeps an incorrect final answer.}
\label{fig:rollout_trace_repair_example}
\vspace{-1.0ex}
\end{figure*}

\subsection{Illustrative Rollout Trace}
\label{app:rollout_trace}

Figures~\ref{fig:rollout_trace_example} and~\ref{fig:rollout_trace_repair_example} give abridged rollout traces for two training examples.
Figure~\ref{fig:rollout_trace_example} shows a correct candidate that receives Privileged Feedback confirmation during training.
Figure~\ref{fig:rollout_trace_repair_example} shows the main repair case, where an initially wrong candidate is corrected after Privileged Feedback.
The traces omit policy reasoning text, normalize tool names to the notation used in this paper, and separate policy-visible Privileged Feedback calls from the hidden verifier-side fields passed to the feedback generator.
Reference answers are shown only to identify hidden fields for these examples and are not part of the policy-visible context.

\section{Released Artifacts}
\label{app:artifacts}

We will release the code for CAPF training and deployment evaluation, together with the main trained checkpoints.
The training and evaluation datasets are public resources and should be obtained from their original providers.

\section{Licenses}
\label{app:licenses}

The Qwen checkpoints from the Qwen2.5 and Qwen3 model families~\cite{qwen2024qwen25,qwen2025qwen3}, OpenRLHF~\cite{hu2024openrlhf}, and FlashRAG~\cite{jin2024flashrag} should be used under their upstream licenses.

\end{document}